\documentclass[10pt,twocolumn,letterpaper]{article}

\usepackage[pagenumbers]{cvpr} % To force page numbers, e.g. for an arXiv version

\definecolor{cvprblue}{rgb}{0.21,0.49,0.74}
\usepackage[pagebackref,breaklinks,colorlinks,allcolors=cvprblue]{hyperref}

\usepackage{graphicx}
\usepackage{amsmath}
\usepackage{amssymb}
\usepackage{booktabs}

\usepackage{multirow}
\usepackage{bbding}
\usepackage{utfsym}

\usepackage{xspace}
\usepackage{enumitem}

\usepackage{makecell}
\usepackage{pifont}

\usepackage{colortbl}
\usepackage[dvipsnames]{xcolor}

\usepackage[accsupp]{axessibility}  % Improves PDF readability for those with disabilities.

\newcommand{\systemname}{{RaCFormer}\xspace}
\newcommand{\parsection}[1]{\vspace{4pt}\noindent\textbf{#1:}}

\def\confName{CVPR}
\def\confYear{2025}

\title{\systemname: Towards High-Quality 3D Object Detection via Query-based Radar-Camera Fusion}

\author{Xiaomeng Chu$^1$, Jiajun Deng$^{2*}$, Guoliang You$^1$, Yifan Duan$^1$, Houqiang Li$^1$, Yanyong Zhang$^{1,3}$\thanks{Corresponding authors.} 
\\
\\
$^1$ University of Science and Technology of China,
$^2$ The University of Adelaide, \\
$^3$ Institute of Artificial Intelligence, Hefei Comprehensive National Science Center
}

\begin{document}
\maketitle

\begin{abstract}
We propose \textbf{Ra}dar-\textbf{C}amera fusion trans\textbf{former} (\systemname) to boost the accuracy of 3D object detection by the following insight.
The Radar-Camera fusion in outdoor 3D scene perception is capped by the image-to-BEV transformation--if the depth of pixels is not accurately estimated, the naive combination of BEV features actually integrates unaligned visual content.
To avoid this problem, we propose a query-based framework that enables adaptive sampling of instance-relevant features from both the bird's-eye view (BEV) and the original image view. 
Furthermore, we enhance system performance by two key designs: optimizing query initialization and strengthening the representational capacity of BEV.
For the former, we introduce an adaptive circular distribution in polar coordinates to refine the initialization of object queries, allowing for a distance-based adjustment of query density.
For the latter, we initially incorporate a radar-guided depth head to refine the transformation from image view to BEV.
Subsequently, we focus on leveraging the Doppler effect of radar and introduce an implicit dynamic catcher to capture the temporal elements within the BEV.
Extensive experiments on nuScenes and View-of-Delft (VoD) datasets validate the merits of our design. Remarkably, our method achieves superior results of 64.9\% mAP and 70.2\% NDS on nuScenes.
\systemname also secures the state-of-the-art performance on the VoD dataset.
Code is available at \href{https://github.com/cxmomo/RaCFormer}{https://github.com/cxmomo/RaCFormer}.
\end{abstract}
\vspace{-0.2cm}
\section{Introduction}

\begin{figure}[t]
    \centering
    \includegraphics[width=0.93\columnwidth]{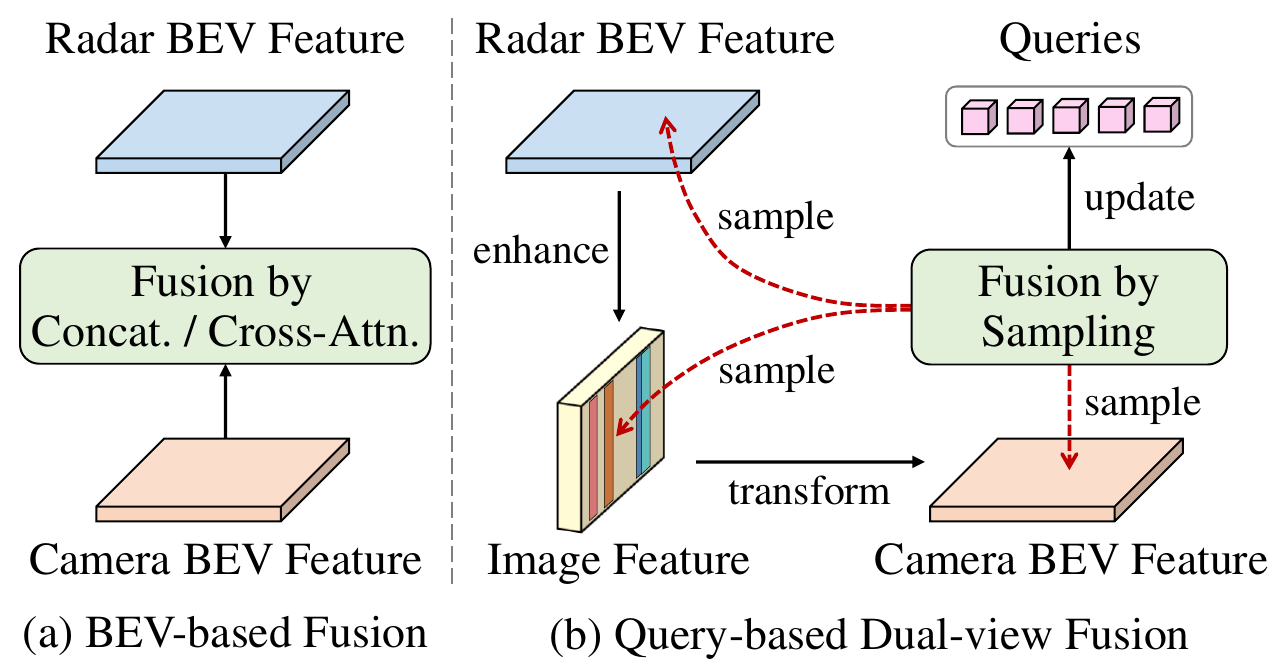}
    \vspace{-0.1cm}
    \caption{ 
    Motivation of \systemname. (a) Previous methods typically fuse BEV features from image-view transformation and radar point cloud encoding, by concatenation or cross-attention. (b) Instead, \systemname uses a query-based fusion framework by simultaneously sampling radar-enhanced image-view features, camera-transformed BEV features, and radar-encoded BEV features.
    }  
    \label{fig:comp}
    \vspace{-0.2cm}
\end{figure}

Precise 3D object detection plays a vital role in promoting the safety and efficiency of autonomous vehicles and intelligent robotic systems~\cite{3dod_survey3, 3dod_survey4, 3dod_tpami_survey, 3dod_ijcv_survey1}.
Compared to adopting the expensive LiDAR sensor, the solution with multi-view cameras and millimeter-wave radar is more applicable due to the dramatically reduced cost, thus attracting a surge of research interests in the community~\cite{RadarNet,CenterFusion}.

Despite impressive advancement, it is still a non-trivial problem in approaching the detection accuracy of LiDAR-based methods by the combination of camera and radar data.
Current top-performing radar-camera fusion approaches~\cite{RCBEV, CRN, RCBEVDet, rcfusion} typically adopt the BEV-based fusion framework~\cite{BEVFusion} which unifies the representation in BEV space to facilitate fusion.
In this paradigm, image and radar features are independently extracted, unified into a BEV representation, and subsequently fused by concatenation or cross-attention, as shown in Fig.~\ref{fig:comp} (a).
However, the inherent disparity between the two modalities poses a significant challenge when relying solely on BEV features for fusion.
Due to radars' hardware constraints, e.g., bandwidth and array design, their limited spatial resolution results in sparse radar BEV features.
On the other hand, camera BEV features are generated from dense image features, but have the issue of feature distortion due to the inaccurate depth estimation in view transformation~\cite{LSS}.
In contrast, the original perspective image features offer a semantically rich representation free from distortions, which has the potential to aid camera-radar fusion.
This underscores the necessity of heterogeneous feature fusion across perspectives of the front view and BEV.
Inspired by this finding, this work aims to explore an effective cross-perspective fusion framework that can accommodate resolution and semantic discrepancies between the two modalities.

Motivated by the above analysis, we ask one essential question: What kind of fusion paradigm can remain unaffected by feature density while effectively utilizing information from different views? Back to the detection algorithms, the query-based approach~\cite{DETR} shows potential in addressing this dilemma. Specifically, the object query initialized in the 3D space can be leveraged as a medium for abstracting features from arbitrary projection views.

Formally, we propose \systemname, a query-based radar-camera fusion framework that improves camera-radar fusion by sampling object-relevant features from both perspective and bird's-eye views, as illustrated in the dual-view fusion paradigm in Fig.~\ref{fig:comp} (b). Our framework has three main designs: linearly increasing circular query initialization, radar-aware depth prediction, and temporal radar BEV encoding. 
The first optimizes the initialization distribution of queries, while the latter two refine and bolster the BEV representation.
Specifically, we propose a circular query initialization strategy that places query points along concentric circles to align the projection principle of sensors.
Additionally, we ensure a linear increase in the number of queries from inner to outer circles, thereby mitigating the issue of queries being much sparser at distant ranges compared to nearby areas.
Furthermore, conventional automotive radars exhibit significant height estimation errors due to their limited vertical angular resolution.
Therefore, we assign a default height and project radar points onto the image features to enhance depth prediction for view transformation, refining the camera BEV features.
We also utilize the radar's Doppler effect to track object velocities by employing an implicit dynamic catcher with convolutional gated recurrent units, effectively capturing temporal elements on multi-frame radar BEV features.

To demonstrate the effectiveness of our proposed \systemname, we benchmark our method on the challenging nuScenes~\cite{nuScenes} dataset and the View-of-Delft (VoD)~\cite{vod} dataset. Without bells and whistles, our approach achieves 64.9\% mAP and 70.2\% NDS on the nuScenes test set. 
Remarkably, on the VoD dataset, our method achieves a 54.4\% mAP across the entire annotated area and a 78.6\%  mAP in the region of interest, earning a 1st-ranking performance.

In summary, our main contributions are as follows:
\begin{itemize}
\item We introduce \systemname, an innovative query-based 3D object detection method through cross-perspective radar-camera fusion. Object queries are initialized to a linearly increasing circular distribution, which aligns with camera projection principles and ensures reasonable density.

\item On the image view, we refine depth estimation using the radar-aware depth head, facilitating more accurate transformations from the image plane to the BEV. Concurrently, on the BEV, we bolster the motion perception of radar BEV features with the implicit dynamic catcher.

\item We perform experiments on the nuScenes and VoD datasets. Our method achieves state-of-the-art performance on both the nuScenes test set and the VoD dataset. 

\end{itemize}

\section{Related Work}

\parsection{Camera-based 3D Object Detection}
Multi-camera 3D object detection methods fall mainly into BEV-based and query-based categories.
Notable BEV-based approaches, such as BEVDet~\cite{BEVDet} and BEVDepth~\cite{BEVDepth}, apply the lift-splat-shoot method~\cite{LSS} to transform the image view into a top-down perspective. 
BEVFormer~\cite{BEVFormer} uses deformable cross-attention for the construction of BEV features and integrates temporal data. FB-BEV~\cite{FB-BEV} improves BEV representations with a forward-backward view transformation module, while HOP~\cite{HOP} employs temporal decoders to predict objects using pseudo-BEV features to capture dynamics. VideoBEV~\cite{videobev} stands out with its long-term recurrent fusion technique, seamlessly incorporating historical data.
On the other hand, query-based methods such as DETR3D~\cite{DETR3D} and PETR~\cite{PETR} harness the transformer decoder to interpret image features. 
StreamPETR~\cite{streampetr} extends PETR with an object-centric temporal mechanism for long-sequence modeling, using frame-by-frame object query propagation. 
MV2D~\cite{mv2d} enhances detection capabilities by using 2D detectors to generate object-specific queries, while RayDN~\cite{RayDN} improves detection precision by strategically sampling camera rays to generate depth-aware features.
Sparse4D~\cite{sparse4D} refines the anchor boxes through sparse feature sampling, assigning multiple 4D key points to each 3D anchor.
Lastly, SparseBEV~\cite{sparsebev} introduces a fully sparse 3D detection framework, fusing scale-adaptive attention with adaptive spatio-temporal sampling.

\begin{figure*}[t]
    \centering
    \includegraphics[width=0.95\textwidth]{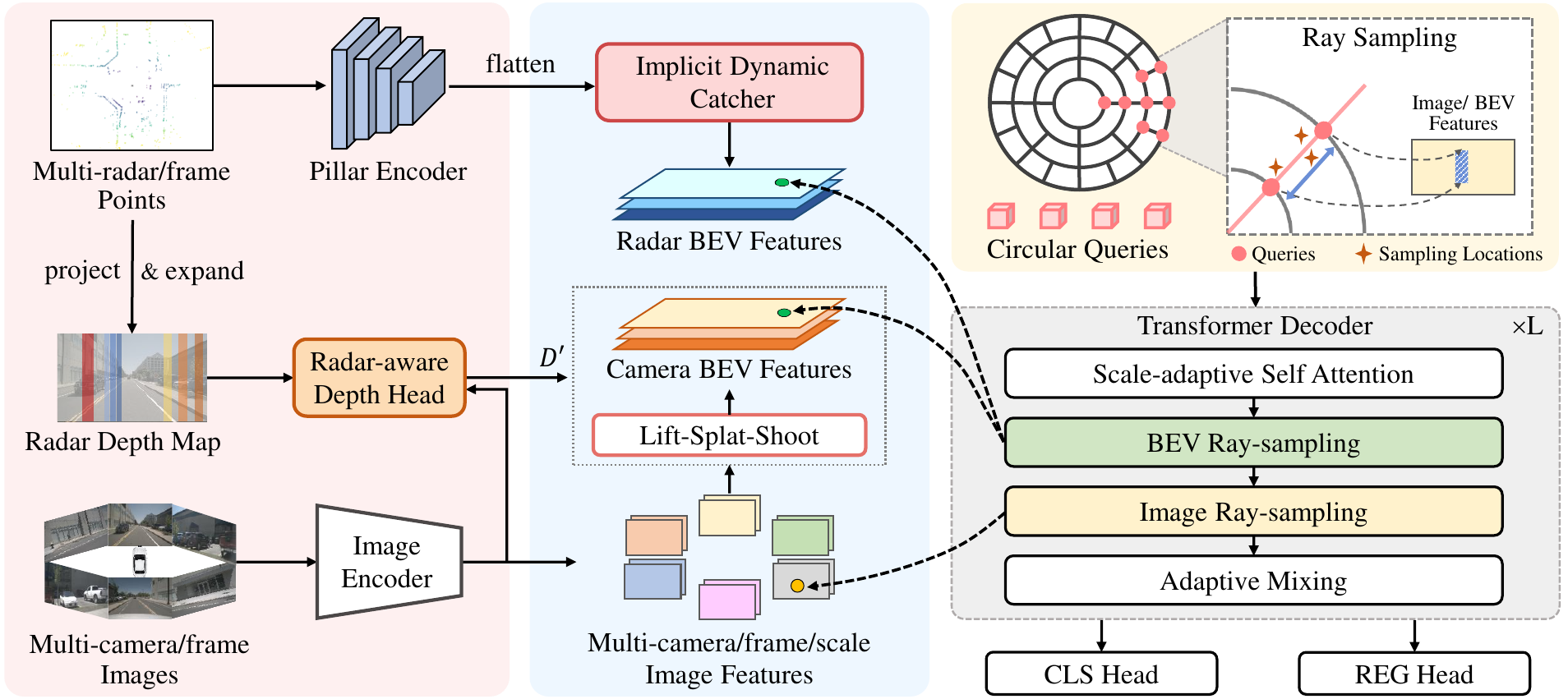}
    \vspace{-0.1cm}
    \caption{
    Overall architecture of \systemname. The image encoder extracts features from multiple frames of multi-camera images, while multi-frame radar points are voxelized and processed by a pillar encoder. The radar features are flattened into the BEV and enhanced by an implicit dynamic catcher. Simultaneously, radar points are re-projected onto the image plane, with their depth values extended to the full image height, and merged with image features in the depth head to refine depth prediction. The refined depth probability distribution $D'$ and the image features are then input into the lift-splat-shoot (LSS) module to create camera BEV features. The transformer decoder initializes queries with an adjustable circular distribution. Over $L$ layers, a ray sampling module within each layer extracts both image-view and BEV features to refine queries, enabling precise classification and regression by the subsequent heads.
    }  
    \label{fig:pipeline}
    \vspace{-0.2cm}
\end{figure*}

\parsection{Radar-Camera Fusion-based 3D Object Detection}
In pursuit of precise 3D object detection, various innovative sensor fusion techniques have emerged~\cite{TransFusion, cmt, PoIFusion, IS-Fusion, EZFusion, rcfusion}, with radar-camera fusion gaining widespread research attention.
CRN~\cite{CRN} generates a detailed BEV feature map by integrating camera and radar data, transforming image features into BEV, and applying multi-modal deformable attention to resolve spatial misalignment.
Subsequently, HVDetFusion~\cite{HVDetFusion} accommodates both camera-only and radar-camera inputs, enhancing BEVDet4D~\cite{BEVDet4D} for camera streams and refining radar data with object priors to supplement and fuse the BEV features.
CRAFT~\cite{CRAFT} proposes an early fusion strategy at the proposal level, combining spatial and contextual data from cameras and radars. 
Meanwhile, RADIANT~\cite{RADIANT} corrects monocular depth errors by predicting 3D offsets between radar returns and object centers, improving accuracy without retraining existing models.
RCBEVDet~\cite{RCBEVDet} introduces RadarBEVNet, a pioneering module for radar feature extraction in BEV, coupled with a fusion mechanism that autonomously aligns the multi-modal BEV features
Lastly, HyDRa~\cite{HyDRa} employs a hybrid approach to merge camera and radar features in both perspective and BEV spaces, including a height association transformer for reliable depth estimation.

\section{Method}

\subsection{Overall Framework}

\systemname, as shown in Fig.~\ref{fig:pipeline}, offers a query-based 3D object detection framework that integrates radar and camera inputs. 
The core modules of the framework include an image encoder, a pillar encoder, a radar-aware depth head, an LSS view transformation module, an implicit dynamic catcher, and a transformer decoder. 
The image encoder extracts features from camera frames, while the pillar encoder processes radar points and flattens the features to BEV. 
Subsequently, the radar BEV features are refined by an implicit dynamic catcher to capture moving elements.
Radar points are also projected into the image plane and combined with visual features in the radar-aware depth head to form a depth probability distribution $D'$. 
The enhanced depth distribution is merged with image features in the LSS module to generate camera BEV features.
Queries serve as a medium for cross-perspective and cross-modality feature fusion, initialized in an adjustable circular distribution and refined by the transformer decoder.
The transformer decoder, comprising $L$ layers, includes a scale-adaptive self-attention module~\cite{sparsebev} for dynamically adjusting receptive fields, two ray-sampling modules for extracting BEV and image-view features, and an adaptive mixer~\cite{AdaMixer} for feature aggregation.
Finally, the classification and regression heads interpret the refined queries for accurate object detection.

\subsection{Camera-transformed BEV Feature Generation with Radar-aware Depth Prediction}

\begin{figure}[t]
    \centering
    \footnotesize
    \begin{tabular}{c}
    \includegraphics[width=0.95\columnwidth]{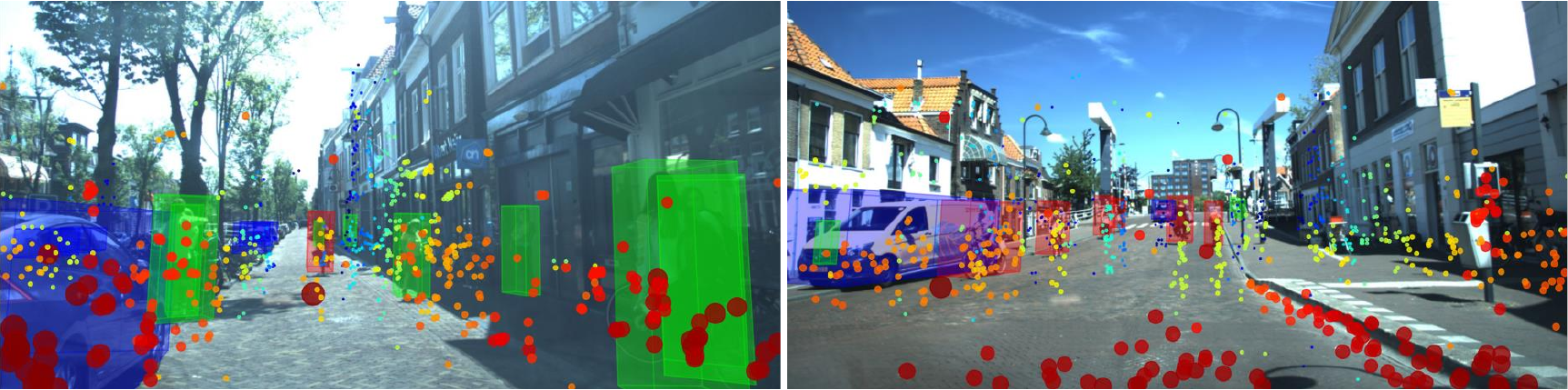} \\
    (a) Radar points projected onto the image \\
    \\
    \includegraphics[width=0.95\columnwidth]{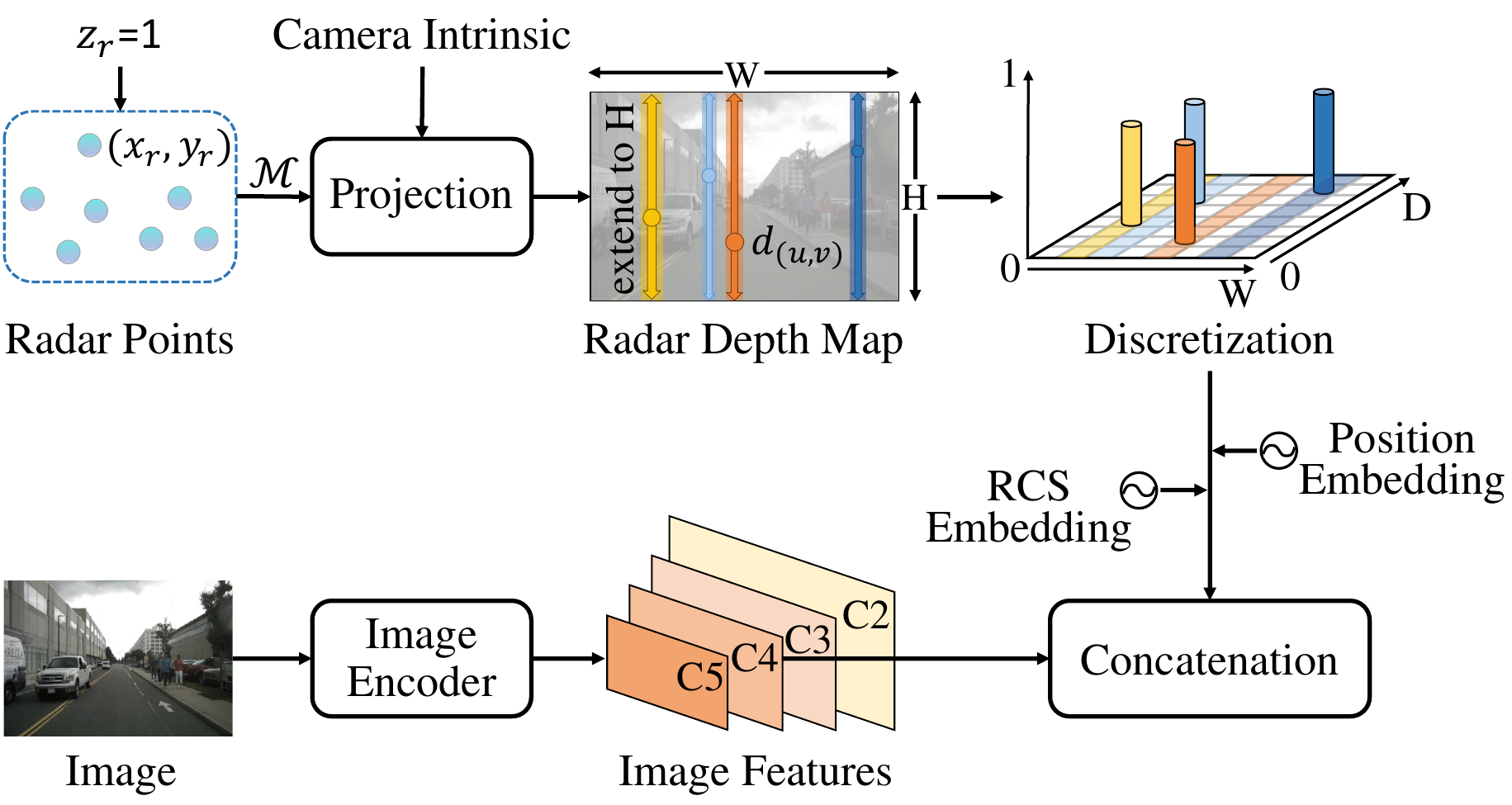} \\
    (b) The pre-processing of the radar-aware depth head \\
    \end{tabular}
    \vspace{-0.1cm}
    \caption{The visualization of (a) radar points with raw z-coordinates projected onto the image and the flowchart of (b) pre-processing input data for the radar-aware depth head.
    }  
    \label{fig:radar-aware_dep}
    \vspace{-0.2cm}
\end{figure}

\parsection{Radar-aware Depth Prediction}
We propose to enhance the image features using radar data for better depth estimation.
However, conventional automotive radars provide distance and velocity measurements within their field of view, but their limited vertical angular resolution leads to significant height estimation errors. 
As depicted in Fig.~\ref{fig:radar-aware_dep}(a), many radar points are projected onto the image with their vertical coordinates falling outside the objects' 2D bounding boxes, due to inaccuracies in the z-coordinates of the raw radar points.
Therefore, we design a pre-processing step before the depth head, shown in Fig.~\ref{fig:radar-aware_dep}(b). 
To maximize the number of radar points falling onto the image's field of view, we initially set $z_r=1$ for all points, denoted as $(x_r, y_r, z_r)$, and then project them onto the image plane using the camera's intrinsic parameters.
The specific transformation formula is as follows:
\begin{equation}
\centering
\begin{aligned}
(x_c, y_c, d) = \mathcal{M} \cdot (x_r&, y_r, z_r),\quad
z_r = 1, \\
u = \frac{f_x \cdot x_c}{d} + c_x,\quad&
v = \frac{f_y \cdot y_c}{d} + c_y,
\end{aligned}
\end{equation}
where $\mathcal{M}$ is the transformation matrix mapping radar coordinates to camera coordinates. The focal lengths $f_x$ and $f_y$ correspond to the camera's x and y axes, respectively, while $c_x$ and $c_y$ specify the image's principal point location.

Next, we extend the vertical coordinate of each projected point to the full height $H$ of the image and assign its depth value, creating a coarse radar depth map. We then use a spacing-increasing discretization strategy~\cite{DORN} to discretize these depths within the range $[0, \text{D}]$.
Furthermore, the Radar Cross Section (RCS) attribute indicates an object's detectability. We embed RCS and the pixel position of the projected radar points into the discretized depth to generate comprehensive radar-aware features, which are concatenated with the $16\times$ downsampled image features $C4$ and input into the depth head.

\parsection{Camera-transformed BEV Feature Generation}
We follow the methods established by BEV-based 3D object detection works~\cite{BEVDet, BEVDepth} and employ the lift-splat-shoot~\cite{LSS} approach for transformation from image view to BEV. 
The process begins with lifting the 2D image features into a 3D space using discretized depth. 
The lifted features are then splatted or distributed onto the BEV plane according to their 3D positions.
The shooting step involves rendering the BEV features for subsequent perception tasks.

\subsection{Radar-encoded BEV Feature Generation with Implicit Dynamic Catcher}

\begin{figure}[t]
    \centering
    \includegraphics[width=0.98\columnwidth]{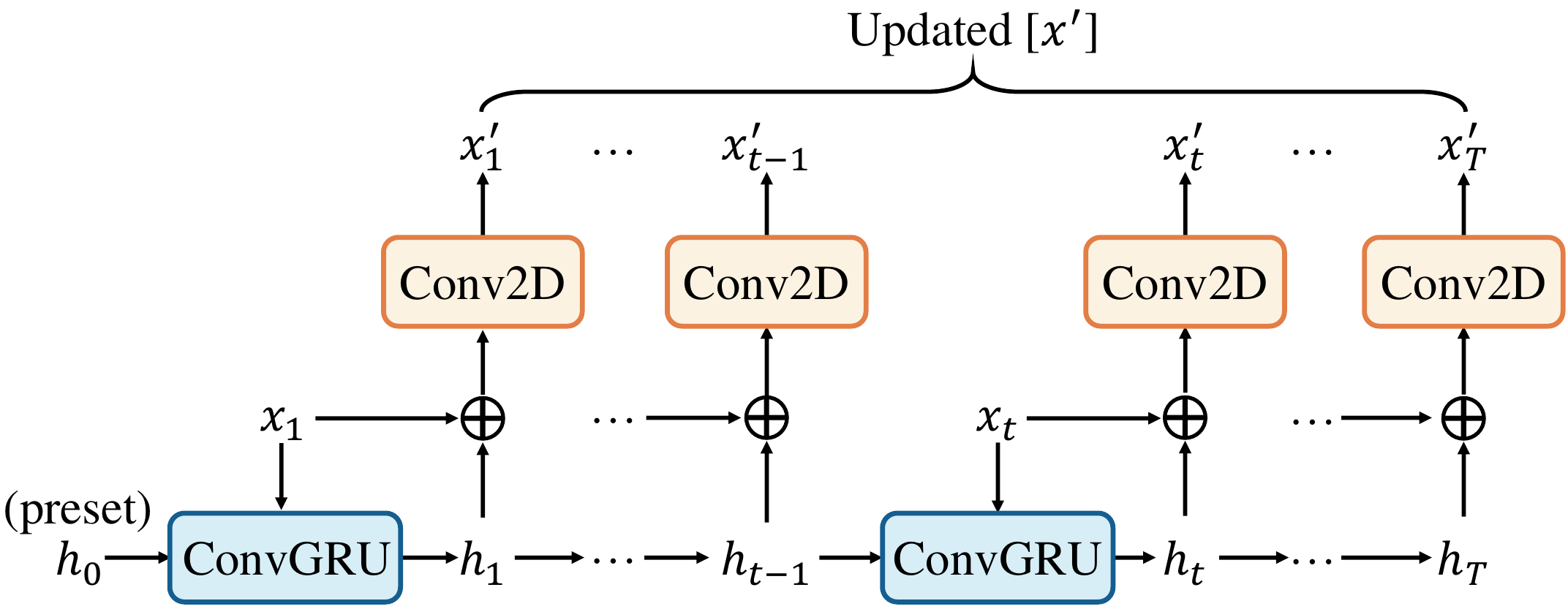}
    \vspace{-0.1cm}
    \caption{The structure of our implicit dynamic catcher. $h_{t}$ represents the hidden state at time $t$, with $h_{0}$ being a preset value of zeros. $x_{t}$ denotes the BEV features output by the pillar encoder at time $t$, while $x'_{t}$ indicates the updated BEV features from $x_{t}$.
    }  
    \label{fig:temp_bev_encoder}
    \vspace{-0.2cm}
\end{figure}

\parsection{Implicit Dynamic Catching}
Millimeter-wave radars leverage the Doppler effect for velocity measurement of moving objects.
To harness this, we introduce an implicit dynamic catcher module designed to capture the temporal elements from multi-frame radar-derived BEV features.
The ConvGRU, an extension of the GRU that integrates convolutional layers, excels at processing sequential data while discerning spatial hierarchies. This makes it an ideal core component for our implicit dynamic catcher, as depicted in Fig.~\ref{fig:temp_bev_encoder}.
Specifically, the dynamic catcher involves accumulating hidden states across consecutive frames $0\sim T$. For instance, the BEV feature of the $t$-th frame, $x_t$, along with the previous frame's hidden state, $h_{t-1}$, are fed into the ConvGRU. This process yields the current frame's hidden state, $h_t$. Subsequently, $h_t$ is combined with $x_t$, and goes through a 2D convolutional layer to produce the refined BEV feature $x'_t$. The process is expressed as follows:
\begin{equation}
\begin{aligned}
h_t &= \text{ConvGRU}(x_t, h_{t-1}), \\
x'_t &= \text{Conv2D}(h_t \oplus x_t).
\end{aligned}
\end{equation}

\parsection{Radar-encoded BEV Feature Generation}
\systemname processes raw radar data by encoding it in a manner analogous to LiDAR point clouds, utilizing a pillar-based method~\cite{PointPillars}.
We begin by setting the $z$-coordinates of radar points to zero and then project them onto the BEV plane using their $(x, y)$ coordinates. The BEV perception range is then segmented into small square pillars, each corresponding to a specific local area.
Within each pillar, we apply a pillar feature network to process the enclosed point cloud data to generate local features. Finally, we construct a BEV feature map by performing max pooling across these pillars.

\subsection{Query Initialization and Ray Sampling}

\parsection{Linearly Increasing Circular Query Distribution}
The radial query initialization from RayFormer~\cite{RayFormer} mimics camera rays, reducing the overlap of different queries projecting on single objects. However, it results in dense queries near the camera and sparser coverage at greater distances, affecting far-object detection. To address this, we introduce a circular query initialization that linearly increases query density according to distance, with adjustable coefficients.

\begin{figure}[t]
    \centering
    \includegraphics[width=0.95\columnwidth]{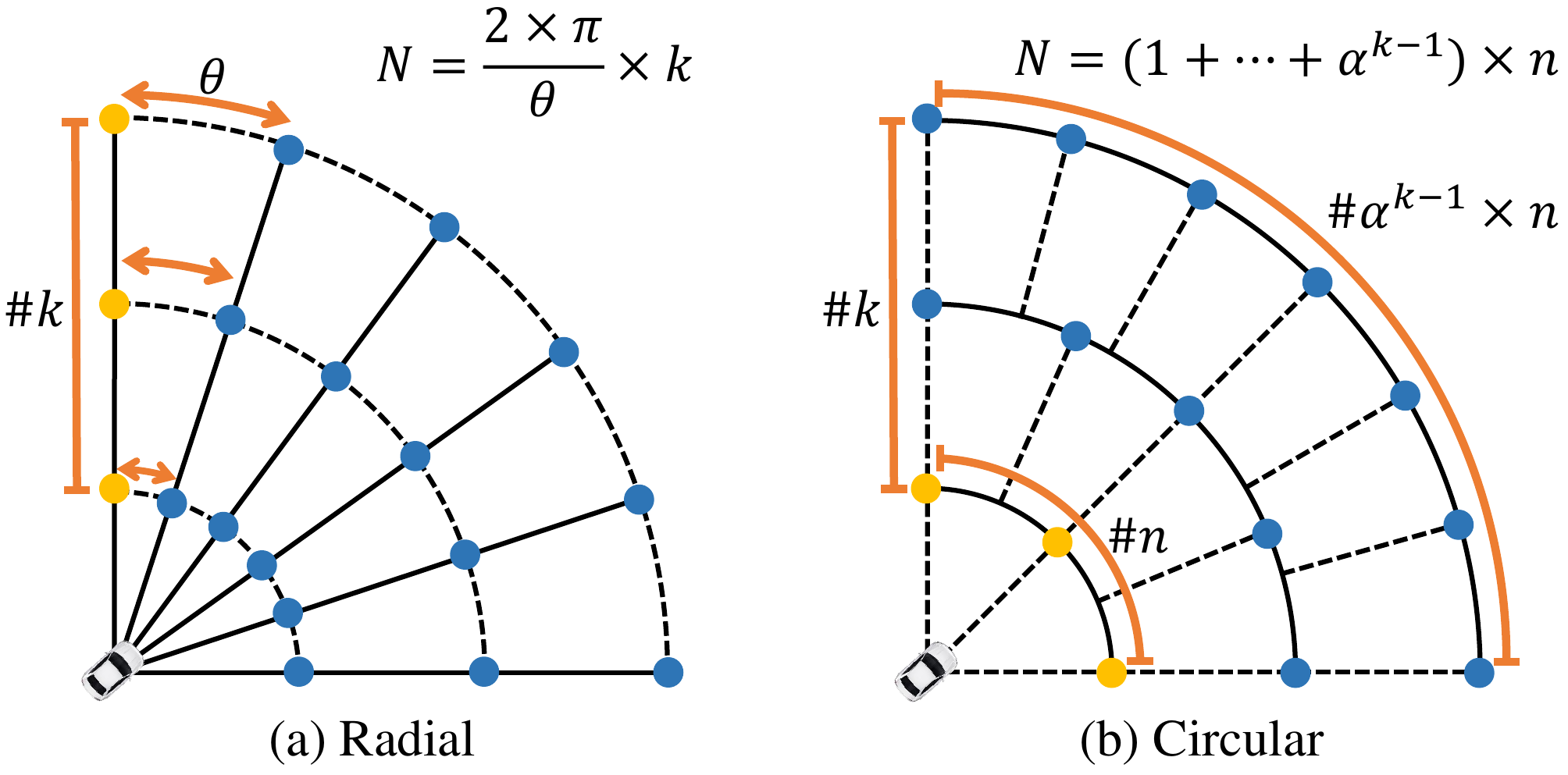}
    \vspace{-0.2cm}
    \caption{
    Comparison of query initialization methods: 
    (a) Radial distribution: Queries are evenly spaced along each ray, with a constant angle $\theta$ separating adjacent rays.
    (b) Linearly increasing circular distribution: The parameter $n$ denotes the query count in the innermost circle, and the linear growth factor of each outer circle is $\alpha$.
    The parameter $k$ indicates the query count per ray in (a) and the number of concentric circles in (b).
    }  
    \label{fig:query_init}
    \vspace{-0.2cm}
\end{figure}

As depicted in Fig.~\ref{fig:query_init}(a), the radial distribution emits rays from the BEV center at uniform angles $\theta$ and places $k$ queries per ray. The total query count $N$ is given by:
\begin{equation}
N = \frac{2\times\pi}{\theta}\times k.
\end{equation}
Our proposed circular initialization method, as illustrated in Fig.~\ref{fig:query_init}(b), organizes queries in $k$ concentric circles. Starting with $n$ queries in the innermost circle, each subsequent circle contains $\alpha$ times the query count of the adjacent inner one, up to $\alpha^{k-1} \times n$ queries in the outermost circle. The total query count $N$ is calculated as follows:
\begin{equation}
\begin{aligned}
N &= (1+\alpha+...+\alpha^{k-1})\times n \\
  &= 
  \begin{cases}
  k\times n, \quad &\alpha = 1, \\
  \frac{\alpha^{k}-1}{\alpha-1}\times n, \quad &\alpha \neq 1. 
  \end{cases}
\end{aligned}
\end{equation}
When $\alpha = 1$, all circles have equal query counts, making the method equivalent to the radial one in this specific case.

\parsection{Ray Sampling Across Perspectives and Modalities}
We employ the ray sampling method following RayFormer~\cite{RayFormer}, which takes ray segments as units, reflecting the natural optical properties of cameras. 
In this approach, each query defines a segment whose length corresponds to the interval between adjacent circles. Within this segment, several adaptive sampling points are selected to gather features from the image view and the BEV.
For BEV ray sampling, we integrate historical BEV features into the ego coordinate system and apply deformable attention. 
Image ray sampling involves projecting sampling points onto multi-camera images from multiple timestamps to extract pixel features.
Finally, the adaptive mixing process~\cite{AdaMixer, sparsebev} aggregates the spatio-temporal features across channels and points.

\section{Experiments}

\subsection{Datasets and Metrics}

\parsection{NuScenes}
NuScenes~\cite{nuScenes}, renowned for its extensive perception challenges, is meticulously divided into 1,000 instances, with 700 for training, 150 for validation, and 150 for testing. Each scene, annotated at 2Hz, provides a 20-second duration. The nuScenes evaluation metrics, including average translation error (ATE), average scale error (ASE), average orientation error (AOE), average velocity error (AVE), and average attribute error (AAE), assess the precision of object detection in terms of position, size, orientation, motion, and attributes. The mean Average Precision (mAP) and the nuScenes detection score (NDS) further measure the overall effectiveness of detection systems.

\parsection{View-of-Delft (VoD)}
VoD~\cite{vod} comprises 8693 frames, each containing synchronized and calibrated 64-layer LiDAR-, (stereo) camera-, and 3+1D radar-data, all captured in complex urban traffic scenarios.
It features 123,106 3D bounding box annotations for a variety of moving and stationary objects, encompassing pedestrian, cyclist, and car. 
The evaluation criteria follow the KITTI~\cite{KITTI1, KITTI2} dataset, which assesses the detection performance using the mean Average Precision metric of 3D bounding boxes.

\begin{table*}[t]
  \centering
  % \small
  \footnotesize
    \setlength\tabcolsep{6pt}
    \begin{tabular}{l|cccc|cc|ccccc}
    \toprule
    Methods & Input &Image Size & Backbone & Epochs & \textbf{mAP}$\uparrow$ & \textbf{NDS}$\uparrow$ & mATE$\downarrow$  & mASE$\downarrow$   & mAOE$\downarrow$  & mAVE$\downarrow$  &  mAAE$\downarrow$  \\
    \midrule
    StreamPETR~\cite{streampetr} & C  &256$\times$704    & ResNet50 & 60 & 45.0    & 55.0    & 0.613 & 0.267 & 0.413 & 0.265 & 0.196     \\
    RayFormer~\cite{RayFormer} & C  &256$\times$704    & ResNet50 & 36 & 45.9 &  55.8  & 0.568  &  0.273 & 0.425  &  0.261 & 0.189     \\
    HVDetFusion~\cite{HVDetFusion} & C+R  & 256$\times$704    & ResNet50 &  24  & 45.1    & 55.7    & 0.557 & 0.527 & 0.270 & 0.473 & 0.212     \\
    RCBEVDet~\cite{RCBEVDet} & C+R  & 256$\times$704    & ResNet50 &  12  & 45.3  & 56.8   & 0.486 & 0.285 & 0.404 & 0.220 & 0.192     \\
    CRN~\cite{CRN} & C+R  & 256$\times$704    & ResNet50 &  24 & 49.0  & 56.0   & 0.487 & 0.277 & 0.542 & 0.344 & 0.197     \\
    HyDRa~\cite{HyDRa} & C+R  & 256$\times$704    & ResNet50 &  20 & 49.4  & 58.5   & 0.463 & 0.268 & 0.478 & 0.227 & 0.182     \\
    \rowcolor[gray]{.92}\textbf{\systemname} (Ours) & C+R &256$\times$704   & ResNet50 & 36 &  \textbf{54.1}  &   \textbf{61.3}  & 0.478  &  0.261 & 0.449  &  0.208 & 0.180 \\
    \midrule
    StreamPETR~\cite{streampetr} & C  &512$\times$1408    & ResNet101 & 60 & 50.4 & 59.2 & 0.569 & 0.262 & 0.315 & 0.257 & 0.199     \\  
    RayFormer~\cite{RayFormer} & C &  512$\times$1408   & ResNet101 & 24 & 51.1 & 59.4 & 0.565 & 0.265 & 0.331 & 0.255 & 0.200   \\  
    CRN~\cite{CRN} & C+R  &512$\times$1408    & ResNet101 &  24 & 52.5 & 59.2 & 0.460 & 0.273 & 0.443 & 0.352 & 0.180     \\
    HyDRa~\cite{HyDRa} & C+R  & 512$\times$1408    & ResNet101 & 20  & 53.6 & 61.7 & 0.416 & 0.264 & 0.407 & 0.231 & 0.186     \\
    \rowcolor[gray]{.92}\textbf{\systemname} (Ours) & C+R  &512$\times$1408   & ResNet101 & 24 & \textbf{57.3} & \textbf{63.0} & 0.476 & 0.261 & 0.428 & 0.213 & 0.183   \\  
    \bottomrule
    \end{tabular}
    \vspace{-0.2cm}
    \caption{Comparison of different methods on the nuScenes \texttt{val} set. `C’ and `R’ represent camera and radar, respectively. 
    }
  \label{tab:nus-val}
\end{table*}

\begin{table*}[t]
  \centering
  \footnotesize
    \setlength\tabcolsep{5pt}
    \begin{tabular}{l|cccc|cc|ccccc}
    \toprule
    Methods & Input &Image Size & Backbone & Epochs & \textbf{mAP}$\uparrow$ & \textbf{NDS}$\uparrow$ & mATE$\downarrow$  & mASE$\downarrow$   & mAOE$\downarrow$  & mAVE$\downarrow$  &  mAAE$\downarrow$  \\
    \midrule
    CenterPoint~\cite{CenterPoint3D} & L & - & VoxelNet & 20 & 60.3 & 67.3 & 0.262 & 0.239 & 0.361 & 0.288 & 0.136 \\
    VoxelNeXt~\cite{VoxelNeXt} & L & - & Sparse CNNs & 20 & 64.5 & 70.0 & 0.268 & 0.238 & 0.377 & 0.219 & 0.127 \\
    \midrule
    UVTR~\cite{UVTR} & C &900$\times$1600   & V2-99 & 24 & 47.2 & 55.1 & 0.577 & 0.253 & 0.391 & 0.508 &  0.123 \\
    PolarFormer~\cite{PolarFormer} & C  &640$\times$1600 & V2-99 & 24 & 49.3 & 57.2 & 0.556 & 0.256 & 0.364 & 0.439 & 0.127    \\
    RayFormer~\cite{RayFormer} & C & 640$\times$1600   & V2-99 & 24 & 55.5  &  63.3 & 0.507 & 0.245 & 0.326 & 0.247 & 0.123   \\  
    RCBEVDet~\cite{RCBEVDet} & C+R  & 640$\times$1600    & V2-99 &  12  & 55.0  & 63.9   & 0.390 & 0.234 & 0.362 & 0.259 & 0.113     \\
    CRN~\cite{CRN} & C+R  &  640$\times$1600    & ConvNeXt-B & 24  & 57.5  & 62.4   & 0.416 & 0.264 & 0.456 & 0.365 & 0.130    \\
    HyDRa~\cite{HyDRa} & C+R  &  640$\times$1600   &  V2-99 & 20  & 57.4  & 64.2   & 0.398 & 0.251 & 0.423 & 0.249 & 0.122     \\
    \rowcolor[gray]{.92}\textbf{\systemname}  (Ours) & C+R  &  640$\times$1600   & V2-99 & 24 & \textbf{59.2} & \textbf{65.9} & 0.407 & 0.244 & 0.345 & 0.238 & 0.132   \\
    \midrule
    HVDetFusion~\cite{HVDetFusion} (+8) & C+R  &  640$\times$1600   &  InternImage-B & 20  & 60.9 & 67.4 & 0.379 & 0.243 & 0.382 & 0.172 & 0.132     \\
    \rowcolor[gray]{.92}\textbf{\systemname} (+6) & C+R  &  640$\times$1600   & V2-99 & 24 & \textbf{64.9} & \textbf{70.2} & 0.358 & 0.240 & 0.329 & 0.179 & 0.119  \\ 
    \bottomrule
    \end{tabular}
    \vspace{-0.2cm}
    \caption{Comparison on the nuScenes \texttt{test} set. 
    The VoVNet-99 (V2-99)~\cite{vovnet} is pre-trained from DD3D~\cite{DD3D} with extra data. 
    `L', `C', and `R' represent LiDAR, camera, and radar, respectively.
    ``(+$t$)" indicates using future and historical frames, each by $t$ frames.
    }
  \label{tab:nus-test}
  \vspace{-0.1cm}
\end{table*}

\subsection{Implementation Details}
For the nuScenes dataset, our BEV perception area, encompassing a 65-meter radius circle, is segmented into $k=6$ concentric circles. We initiate with $n=80$ queries in the innermost circle, expanding outwards by a factor of $\alpha\approx1.25$ per subsequent circle, culminating in 900 queries overall.
Due to the use of monocular images, the perception area of VoD is a 55-meter radius sector spanning $3/4\pi$ radians, divided into $k=8$ concentric arcs. We allocate $n=30$ queries to the inner arc, maintaining the $\alpha\approx1.25$ and totaling 600 queries.
Our transformer decoder comprises 6 layers with shared weights for efficiency. 
Additionally, we adopt the query denoising strategy derived from PETRv2~\cite{petrv2} to accelerate convergence.

Our models are trained using the AdamW~\cite{adamw} optimizer with a global batch size of 8. We initiate the learning process with a learning rate of 2e-5 for the backbone and 2e-4 for other parameters, applying a cosine annealing~\cite{cosinedecay} policy for rate adjustment. For image feature encoding, we adopt the standard networks ResNet~\cite{ResNet} and VoVNet-99 (V2-99)~\cite{V2-99}. In line with established practices~\cite{sparsebev, sparse4D, streampetr}, the ResNet parameters are pre-trained on nuImages~\cite{nuScenes}, and the V2-99 parameters are pre-trained on DD3D~\cite{DD3D} with additional datasets. 
Unless specifically indicated, training is conducted for a standard 24 epochs for all models.

\begin{figure*}[t]
    \centering
    \footnotesize
    \setlength\tabcolsep{1.2pt}
    \begin{tabular}{cc}
    \rotatebox{90}{\;\;\;\;\:\:(a) Rainy} &\includegraphics[width=0.93\linewidth]{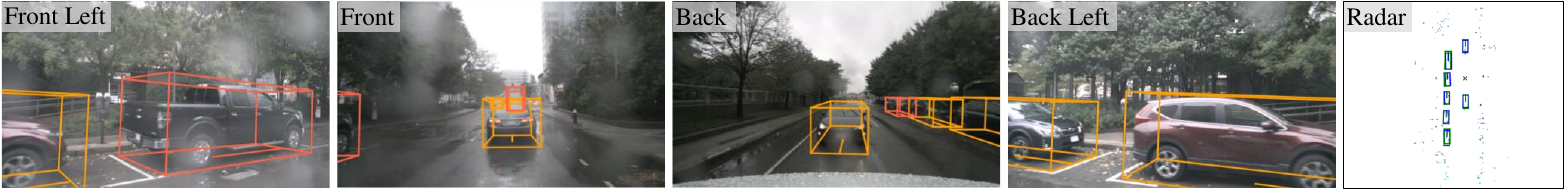} \\
    \rotatebox{90}{\;\;\;\;\:\:(b) Night} & \includegraphics[width=0.93\linewidth]{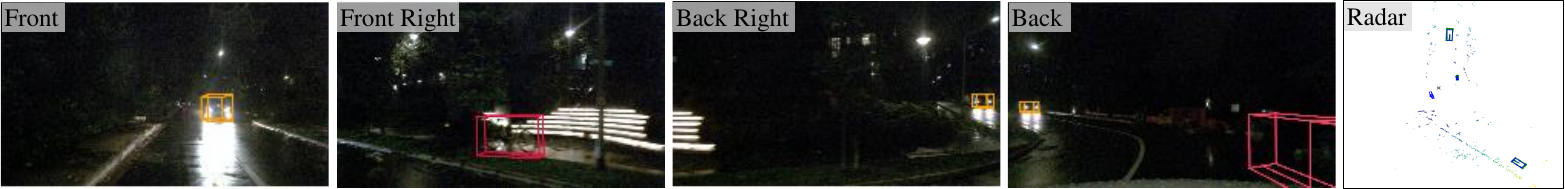} \\
    \rotatebox{90}{\:(c) Object-filled} & \includegraphics[width=0.93\linewidth]{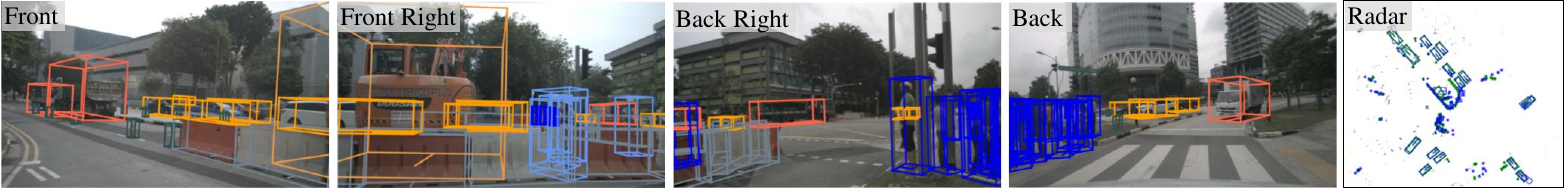} \\
    \end{tabular} 
    \vspace{-0.2cm}
    \caption{Qualitative analysis across varied scenarios—rainy, nighttime, and object-filled. Images (left) exhibit 3D bounding boxes in diverse colors for different categories, while the BEV radar point clouds (right) depict ground truth in \textcolor[RGB]{0,128,0}{green} and predicted boxes in \textcolor[RGB]{14,14,255}{blue}.
    } 
    \label{fig:visual}
\end{figure*}

\begin{table*}[t]
  \centering
  \footnotesize
    \setlength\tabcolsep{11.4pt}
    \begin{tabular}{l|c|ccc|c|ccc|c}
    \toprule
    \multirow{2}{*}{Methods} & \multirow{2}{*}{Input} & \multicolumn{4}{c|}{AP in the Entire Annotated Area (\%)} & \multicolumn{4}{c}{AP in the Region of Interest (\%)} \\
    \cline{3-10}
    & & Car & Pedestrian & Cyclist & mAP & Car & Pedestrian & Cyclist & mAP \\
    \midrule
    
    PointPillars~\cite{PointPillars} & R & 37.06 & 35.04 & 63.44 & 45.18 & 70.15 & 47.22 & 85.07 & 67.48 \\
    RadarPillarNet~\cite{rcfusion} & R & 39.30 & 35.10 & 63.63 & 46.01 & 71.65 & 42.80 & 83.14 & 65.86 \\
    RCFusion~\cite{rcfusion} & C+R & 41.70 & 38.95 & 68.31 & 49.65 & 71.87 & 47.50 & 88.33 & 69.23 \\
    RCBEVDet~\cite{RCBEVDet} & C+R & 40.63 & 38.86 & \textbf{70.48} & 49.99 & 72.48 & 49.89 & 87.01 & 69.80 \\
    \rowcolor[gray]{.92}\textbf{\systemname} (Ours) & C+R & \textbf{47.30} & \textbf{46.21} & 69.80 & \textbf{54.44} & \textbf{89.26} & \textbf{56.78} & \textbf{89.67}  &  \textbf{78.57} \\

    \bottomrule
    \end{tabular}
    \vspace{-0.2cm}
    \caption{Comparison of 3D object detection results on VoD \texttt{val} set. The region of interest is the driving corridor located close to the ego-vehicle. The IoU thresholds for AP are set to 0.5 for cars, 0.25 for pedestrians, and 0.25 for cyclists.}
  \label{tab:vod}
  \vspace{-0.1cm}
\end{table*}

\subsection{Main Results}

\parsection{NuScenes Results}
In Tab.~\ref{tab:nus-val} and~\ref{tab:nus-test}, we compare our method with existing state-of-the-art 3D detection methods on the nuScenes validation and test sets. We include both camera-only and radar-camera fusion algorithms for a thorough comparison.
We default to an 8-frame sequence with 0.5-second intervals for comparable analysis.
On the validation set, \systemname equipped with a ResNet-50 backbone at a resolution of $256 \times 704$ surpasses HyDRa~\cite{HyDRa} by 4.7\% in mAP and 2.8\% in NDS, achieving an mAP of 54.1\% and an NDS of 61.3\%. 
When employing ResNet-101 with input dimensions of $512 \times 1408$, \systemname achieves an mAP of 57.3\% and an NDS of 63.0\%, representing a 3.7\% increase in mAP and a 1.3\% increase in NDS compared to HyDRa.
On the test set, with Vovnet-99 as the backbone and 7 historical frames, our method reaches 59.2\% in mAP and 65.9\% in NDS, marking the corresponding enhancements of 1.8\% and 1.7\% over HyDRa. 
Furthermore, we enhance performance by using 6 past and 6 future frames. This results in a 64.9\% mAP and 70.2\% NDS, outperforming HVDetFusion with more input frames by 4.0\% and 2.8\%, respectively.
Additionally, \systemname outperforms representative LiDAR-based methods like VoxelNeXt~\cite{VoxelNeXt} and CenterPoint~\cite{CenterPoint3D}, partially bridging the modality gap.

\parsection{VoD Results}
We evaluate our method by calculating the 3D AP for cars, pedestrians, and cyclists across two regions: the entire annotated area and the region of interest. The results, as detailed in Tab.~\ref{tab:vod}, show that \systemname notably enhances AP across most categories.
Specifically, across the entire annotated area, \systemname achieves a 4.45\% higher mAP compared to RCBEVDet. In the region of interest, \systemname leads with an mAP of 78.57\%, marking an 8.77\% improvement over RCBEVDet, thus demonstrating the state-of-the-art performance.

\parsection{Visualization Results}
Fig.~\ref{fig:visual} presents the qualitative detection outcomes of \systemname in both image view and BEV, showcasing its robustness across various challenging environments. These include adverse conditions such as rain and darkness, as well as object-filled scenarios.

\subsection{Ablation Studies}
Unless specified, we perform ablation studies using single-frame inputs with an image resolution of $256 \times 704$ and a ResNet-50 backbone. For a comprehensive comparison, we benchmark our model against two single-modality detectors: RayFormer~\cite{RayFormer} for camera-based detection and CenterPoint~\cite{CenterPoint3D} for point-based detection.

\parsection{Feature Decoding and Fusion}
Tab.~\ref{tab:ablan_0} illustrates the impact of feature decoding methods and perspective selection. 
The first two rows present the results of employing the standard BEVFusion paradigm~\cite{BEVFusion} using the concatenation operation or deformable cross-attention for BEV feature fusion, combined with a center-head decoder.
Switching to a transformer decoder with queries, focusing solely on the sampling of BEV features, yields a 1.9\% enhancement in mAP and a 2.5\% improvement in NDS, along with a significant reduction in mAVE but an increase in mATE. 
Further sampling of image-view features improves 3.1\% at mAP and 4.1\% at NDS, respectively.

\parsection{Radar-aware Depth Prediction}
In Tab.~\ref{tab:RPE}, we examine the impact of embedding radar points' depth and RCS value. 
The model that only incorporates radar depth embedding achieves an improvement of 0.7\% in mAP and 1.2\% in NDS. 
Similarly, the embedding of RCS enhances 0.5\% mAP and 0.8\% NDS. When both embeddings are used, the performance is further improved, with overall gains of 1.1\% in mAP and 2.1\% in NDS.

\begin{table}[t]
  \centering
  \footnotesize
    \setlength\tabcolsep{4.5pt}
    \begin{tabular}{l|c|cc|ccc}
    \toprule
     Fusion & Views & mAP$\uparrow$ & NDS$\uparrow$ & mATE$\downarrow$ & mASE$\downarrow$ & mAVE$\downarrow$ \\
    \midrule
    Concat. & B  &  35.9  &  42.7  &  0.625 &  0.289  &  0.613 \\
    Def. Attn. & B  & 38.6  &  45.1  & 0.576  &  0.280 &  0.601  \\
    Queries & B  & 40.5  &  47.6  & 0.655 &  0.281 &  0.365  \\
    Queries & B+I &  43.6  &  51.7  &  0.577  & 0.274  &  0.341 \\
    \bottomrule
    \end{tabular}
    \vspace{-0.2cm}
    \caption{Ablation study of the feature decoding. 
    `B' and `I' correspond to the BEV and image view, respectively.}
  \vspace{-0.1cm}
  \label{tab:ablan_0}
\end{table}

\parsection{Implicit Dynamic Catching}
In Tab.~\ref{tab:IDC}, we evaluate the performance of our implicit dynamic catcher (IDC) by evaluating mAP and mAVE for moving objects in the nuScenes validation set, divided by objects' velocity: static (0 m/s), slow ($<$5 m/s), and fast ($>$5 m/s) objects. 
Adding radar data without IDC substantially increases the relative mAP of slow and fast objects by 49.2\% and 48.1\%, while decreasing relative mAVE by 30.2\% and 11.8\%. 
With IDC, \systemname further improves relative mAP by 4.3\% and 3.5\% and relatively reduces mAVE by 2.9\% and 2.0\%. 
The integration of radar data and the IDC module reveals subtle temporal feature changes, leading to more pronounced optimization on slow-moving objects.

\parsection{Linearly Increasing Circular Query Initialization}
In Tab.~\ref{tab:init_hp}, we evaluate the impact of hyper-parameters of circular distribution while keeping the total query count around 900. 
We first initialize the queries using the conventional grid distribution as a baseline (ID=A), then vary the linear growth factor $\alpha$ and the number of circles $k$ to study their effects.
Setting $\alpha$ to 1 and $k$ to 6 (ID=B) matches the radial initialization of RayFormer~\cite{RayFormer}, improving 1.5\% mAP compared to the grid distribution. 
Both $\alpha=2$ (ID=C) and $k=4$ (ID=E) lead to a performance dip, due to unreasonable initial query distribution of over-density in outer circles and low-density in the depth direction, respectively. 
Optimal performance is found with $\alpha$ at 1.25 (ID=D\&F). We adopt $k=6$ as our standard in this paper.

\begin{table}[t]
  \centering
  \footnotesize
    \setlength\tabcolsep{5.9pt}
    \begin{tabular}{cc|cc|ccc}
    \toprule
    Depth & RCS & mAP$\uparrow$ & NDS$\uparrow$ & mATE$\downarrow$ & mASE$\downarrow$ & mAVE$\downarrow$ \\
    \midrule
       &   & 43.6  & 51.7  &  0.577 & 0.274 & 0.341 \\
      \checkmark &   & 44.3  & 52.9  &  0.560 & 0.268 & 0.319  \\
    & \checkmark & 44.1  & 52.5  &  0.564 & 0.269 & 0.332 \\
     \checkmark & \checkmark & 44.7  & 53.8 & 0.558 & 0.270 & 0.313 \\
    \bottomrule
    \end{tabular}
    \vspace{-0.2cm}
    \caption{Ablation study about the depth and RCS embedding in the pre-processing step of radar-aware depth prediction. 
    }
    \vspace{-0.1cm}
  \label{tab:RPE}
\end{table}

\begin{table}[t]
  \centering
  \footnotesize
  \setlength\tabcolsep{7.9pt}
    \begin{tabular}{c|c|c|cc|cc}
    \toprule
    \multirow{2}{*}{Input} & \multirow{2}{*}{IDC} &  0 m/s & \multicolumn{2}{c|}{(0, 5] m/s} & \multicolumn{2}{c}{$>$5 m/s} \\
    \cline{3-7}
    & & mAP & mAP & mAVE & mAP& mAVE \\
    \midrule
    C & \ding{55} & 45.5 & 12.6  & 0.731  &  29.3  & 0.798    \\
    C+R & \ding{55} &  49.4   &  18.8  & 0.510 &  43.4  & 0.704  \\
    C+R & \checkmark & 50.1 & 19.6  & 0.495  &  44.9  & 0.690  \\
    \bottomrule
    \end{tabular}
  \vspace{-0.2cm}
  \caption{
  Analysis of the implicit dynamic catcher (IDC) in detecting moving objects with 8-frame inputs.
}
  \vspace{-0.1cm}
  \label{tab:IDC}
\end{table}

\begin{table}[t]
  \centering
  \footnotesize
    \setlength\tabcolsep{6.5pt}
    \begin{tabular}{l|c|c|c|cc|cc}
    \toprule
    ID & $\alpha$ & k & n & mAP$\uparrow$ & NDS$\uparrow$ & mATE$\downarrow$ & mAVE$\downarrow$  \\
    \midrule
    A  & - & - & - & 42.4 &  52.4 & 0.582  & 0.329  \\
    \cline{2-8}
    B  & 1 & \multirow{3}{*}{6} & 150 & 43.9 &  53.2 & 0.557  & 0.313  \\
    C  & 2 &   & 15 &  43.5 &  52.4 &  0.576 & 0.329  \\
    \cline{2-2}
    D  & \multirow{3}{*}{1.25} &   & 80 & 44.7  & \textbf{53.8} & 0.558 & 0.313 \\
    \cline{3-3}
    E  &   & 4 & 155 & 44.1 &  52.6 &  0.569 & 0.337 \\
    F  &   & 8 & 45 & \textbf{44.8} &  53.6 &  0.549 & 0.339 \\
    \bottomrule
    \end{tabular}
    \vspace{-0.2cm}
    \caption{
    Ablation study on circular query initialization, varying the linear growth factor $\alpha$, the number of concentric circles $k$, and the query count $n$ in the innermost circle.}
   \vspace{-0.1cm}
  \label{tab:init_hp}
\end{table}

\parsection{Weather and Light Conditions}
To verify the impact of radar fusion, we categorize the nuScenes validation set into four scenarios based on weather and lighting: sunny, rainy, daytime, and nighttime. 
We compared our method with two benchmarks--CenterPoint utilizing LiDAR- and RayFormer utilizing camera-input.
As shown in Tab.~\ref{tab:condi}, \systemname surpasses both baselines in all scenarios. Specifically, our method achieves an 8.5\% higher mAP in sunny conditions and a 6.6\% increase in rainy conditions compared to RayFormer. Similarly, it outperforms RayFormer by 8.4\% during the day and by 4.3\% at night.
\systemname reaches 94\% and 89\% of the mAP achieved by LiDAR-based CenterPoint under rain and darkness, respectively, partially compensating for the limitations of image-based perception.

\parsection{Robustness}
To evaluate the robustness gains from radar-camera fusion in our system, we test its performance under sensor failure scenarios, as detailed in Tab.~\ref{tab:robustness}. We systematically exclude either image or radar inputs, recording the AP for detecting cars.
Our method shows a 2.7\% improvement over the CRN under full sensor data conditions and maintains better performance when camera or radar data is missing. Specifically, without any camera data, \systemname still achieves a car AP of 27.2\%, outperforming CRN by 14.4\%. In the absence of radar data, \systemname's car AP is 52.8\%, which is a 9.0\% enhancement over CRN.

\begin{table}[t]
  \centering
  \footnotesize
    \setlength{\tabcolsep}{7.8pt}
    \begin{tabular}{l|c|cccc}
    \toprule
    Methods & Input & Sunny & Rainy & Day & Night  \\
    \midrule
    CenterPoint~\cite{CenterPoint3D} & L & 62.9 & 59.2 & 62.8 & 35.4 \\
    \midrule
    RayFormer~\cite{RayFormer} & C & 44.7  &  49.1  & 45.7  & 27.3 \\
    \rowcolor[gray]{.92}\systemname & C+R & \textbf{53.2} & \textbf{55.7} &  \textbf{54.1} & \textbf{31.6} \\
    \bottomrule
    \end{tabular}
    \vspace{-0.2cm}
  \caption{Analysis of the performance under various weather and lighting scenarios using the mAP metric.}
  \label{tab:condi}
\end{table}

\begin{table}[t]
  \centering
  \footnotesize
  \setlength\tabcolsep{6.6pt}
    \begin{tabular}{l|cc|cccc}
    \toprule
    \multirow{2}{*}{Methods} & \multirow{2}{*}{Input} & \multirow{2}{*}{Drop} & \multicolumn{4}{c}{\# of view drops} \\
    & & & 0 & 1 & 3 & All \\
    \midrule
    CenterPoint~\cite{CenterPoint3D} & R & R & 30.6 & 25.3 & 14.9 & 0 \\
    RayFormer~\cite{RayFormer} & C & C & 52.0 & 44.4 & 24.0 & 0 \\
    \midrule
    \multirow{2}{*}{CRN~\cite{CRN}} & \multirow{2}{*}{C+R} & C & \multirow{2}{*}{68.8} & 62.4 & 48.9 & 12.8 \\
    & & R & & 64.3 & 57.0 & 43.8 \\
    \midrule
    \rowcolor[gray]{.92} &  & C &  &  \textbf{65.6} & \textbf{53.7} & \textbf{27.2} \\
    \rowcolor[gray]{.92}\multirow{-2}{*}{\systemname} & \multirow{-2}{*}{C+R} & R & \multirow{-2}{*}{\textbf{71.5}} & \textbf{69.3} & \textbf{62.5}  & \textbf{52.8} \\
    \bottomrule
    \end{tabular}
    \vspace{-0.2cm}
    \caption{Analysis of robustness using Car class AP. ``All" denotes that the single modality is entirely off.}
  \label{tab:robustness}
  \vspace{-0.1cm}
\end{table}

\parsection{Inference Time}
To enable real-time detection, we develop a lightweight version of our model: it uses 4 historical frames, 450 queries, and a BEV grid of $64\times64$ with a resolution of 1.6 meters per voxel. These adjustments reduce the computational load compared to our standard model. Despite these reductions, our model still delivers state-of-the-art mAP and NDS of 51.0\% and 58.8\%, respectively, outperforming HyDRa~\cite{HyDRa} by 1.6\% in mAP and 2.8\% in NDS. Operating on a single RTX 3090 GPU, it achieves a frame rate of 12 FPS, satisfying real-time requirements.

\section{Conclusion}

In this paper, we present \systemname, a novel query-based 3D object detection method that fuses radar and camera data by cross-perspective feature sampling. 
In particular, by enhancing depth estimation with a radar-guided pre-processing, designing a circular query initialization with the linearly increasing strategy, and leveraging the radar's Doppler effect for BEV temporal encoding, \systemname enables both modalities to capitalize their respective strengths and complement each other effectively.
Our method achieves superior results on the nuScenes and VoD datasets, marking a significant leap forward in high-performance and robust 3D perception for autonomous driving.

\section*{Acknowledgments}
This work was supported by the National Natural Science Foundation of China (No. 62332016) and the Key Research Program of Frontier Sciences, CAS (No. ZDBS-LY-JSC001).

{
    \small
    \bibliographystyle{ieee_fullname}
    \bibliography{main}
}


\end{document}